\documentclass[runningheads]{llncs}

\usepackage{amsmath,amssymb,amsfonts}
\usepackage{yfonts}
\usepackage{hyperref}
\usepackage[pdftex]{graphicx}
\usepackage{textcomp}
\usepackage{xcolor}
\usepackage{multirow}
\usepackage[hang,flushmargin]{footmisc}
\usepackage{lscape}
\usepackage{booktabs}
\usepackage{cite}
\usepackage{subfig}

\begin{document}

\title{A Comprehensive Survey on Outlying Aspect Mining Methods}

\author{Durgesh Samariya \inst{1} \and
Jiangang Ma \inst{1} \and
Sunil Aryal \inst{2}
}

\authorrunning{D. Samariya et al.}

\institute{
School of Science, Engineering and Information Technology, Federation University, Churchill, VIC, Australia \\ \email{\{d.samariya,j.ma\}@federation.edu.au}\\ \and
School of Information Technology, Deakin University, Geelong, VIC, Australia \\
\email{sunil.aryal@deakin.edu.au}}
\maketitle

\begin{abstract}
In recent years, researchers have become increasingly interested in outlying aspect mining. Outlying aspect mining is the task of finding a set of feature(s), where a given data object is different from the rest of the data objects. Remarkably few studies have been designed to address the problem of outlying aspect mining; therefore, little is known about outlying aspect mining approaches and their strengths and weaknesses among researchers. In this work, we have grouped existing outlying aspect mining approaches in three different categories. For each category, we have provided existing work that falls in that category and then provided their strengths and weaknesses in those categories. We also offer time complexity comparison of the current techniques since it is a crucial issue in the real-world scenario. The motive behind this paper is to give a better understanding of the existing outlying aspect mining techniques and how these techniques have been developed.


\end{abstract}

\section{Introduction}
The concept of outliers has been studied extensively in the statistics community from the $19^{th}$ century \cite{Edgeworth1887}. 
In real-world application scenarios, usually there is outlier data, a.k.a. anomaly, which differs from the rest of the data. The word outlier stands for \textit{a statistical observation that is markedly different in value from the others of the sample.}\footnote{https://www.merriam-webster.com/dictionary/outlier. Accessed: 06 April 2020}

Barnett and Lewis(1984) \cite{Barnett1984} formally defined outlier as: ``An observation (or a subset of observations) which appears to be inconsistent with the remainder of that set of data''. \textit{Outlier Detection} (OD) is an essential task in data mining that deals with detecting outliers in data sets automatically. Over the years, an enormous amount of research has been carried out in an attempt to detect outliers in a data set. Although those algorithms are detecting outliers very well, they are not able to explain why those points are considered as an outlier, i.e., they cannot tell in which feature subset(s) the data object significantly deviates from the rest of the data. 

The explanation of outlier has led to a renewed interest in \textit{Outlying Aspect Mining} (OAM). An outlying aspect mining is formally defined as the task of recognizing that feature subset(s) where a given data object is inconsistent with the remainder of that set of data objects. That given data object is called as a query, and those feature subset(s) are called as outlying aspects of the given query. 

The following are some of the definitions found in the literature:

\begin{itemize}
   \item ``\textit{Outlying aspects mining discovers feature subsets (or subspaces) that describe how a query stand out from a given data set.}'' \cite{Wells2019}
   \item \cite{Vinh2016} define outlying aspect mining as ``\textit{problem of investigating, for a particular query object, the sets of features (a.k.a attributes, dimensions) that make it most unusual compared to the rest of the data.}''
\end{itemize}

Previous studies have termed this problem as \textit{outlier explanation} \cite{Micenkova2013}, \textit{outlier interpretation} \cite{Dang2014}, \textit{outlying subspace detection} \cite{Zhang2004}, \textit{outlying aspect mining} \cite{Duan2015, Vinh2016}. A recent line of research has established this problem as outlying aspect mining \cite{Duan2015, Vinh2016, Wells2019, Samariya2020}.

Past studies have hinted at a link between OAM and OD. However, it is worth noting that OAM and OD are different --- the main aim of OAM is to find aspects for a given data object, where it exhibits the most outlying characteristics while the latter focuses on detecting all instances exhibiting outlying characteristics in the given original input space.

Outlying aspect mining has many practical applications, such as an insurance analyst may be interested to find out in which particular aspect an insurance claim looks suspicious. Furthermore, when evaluating job applications, a selection panel wants to investigate in which specific aspect applicant is most different than others. For example, with similar qualifications and experience, John has the highest number of projects completed successfully.

Outlying aspect mining is a new and interesting topic among researchers. To the best of our knowledge, there is no such survey article that has been conducted as of now, which motivates us to write this survey. In this survey paper, we are providing a structured and in-depth review of research on OAM techniques. The work on OAM is categorized into three categories --- 1) Score-and-Search based approach, 2) Feature selection based approach, and 3) Hybrid approach. 

This paper is organized into seven distinct sections. Section \ref{sec:overview} provides an overview of OAM approaches. Outlying aspect mining techniques are categorized in score-and-search based approaches (Section \ref{sec:score-and-search}), feature selection based approaches (Section \ref{sec:feature-selection}) and hybrid approaches (Section \ref{sec:hybrid}). We have discussed open challenges in Section \ref{sec:open-challenges}. Concluding remarks are provided in Section \ref{sec:Conclusion}.

\section{Overview of OAM approaches} \label{sec:overview}

\begin{table}[bt]
    \centering
    \caption{Key symbols and notations used in this paper. }
    \begin{tabular}{l @{\hspace{15pt}} l}
    \toprule\noalign{\smallskip}
    Symbol & Definition \\
    
    \noalign{\smallskip}\midrule\noalign{\smallskip}
    $\mathcal{O}$ & A set of $n$ data instances in an $D$-dimensional space, $|\mathcal{O}|=n$ \\
    ${\bf o}\in \mathcal{O}$ & A data instance represented as a vector, ${\bf o}=\langle o.1,o.2,\cdots,o.D \rangle$ \\
    $\mathcal{F}$ & The set of input features, i.e., $\mathcal{F} = \{1, 2, \cdots, D\}$ \\
    $\mathcal{S}_{\mathcal{F}}$ & The set of all possible subspaces (non-empty subsets) of $\mathcal{F}$ \\
    $d_S({\bf a}, {\bf b})$ & The euclidean distance between ${\bf a}$ and ${\bf b}$ in subspace $S\in \mathcal{S}_{\mathcal{F}}$ \\
    $\aleph_S^k({\bf q})$ & The set of $k$-nearest neighbors of ${\bf q}$ in subspace $S\in \mathcal{S}_{\mathcal{F}}$ \\
    \noalign{\smallskip}\bottomrule
    \end{tabular}
    \label{tab:symbols}
\end{table}

To start with, we have fixed some notations for the rest of the paper and introduced few preliminary definitions. The primary symbols and notations used are provided in Table \ref{tab:symbols}. Let $\mathcal{O} = \{o_1, o_2, \cdots, o_n\}$ be a collection of $n$ data objects in $D$-dimensional space. Each data object ${\bf o}$ is represented as $D$-dimensional vector $\langle o.1, o.2, \cdots, o.D\rangle$. 

As mentioned above, the OAM approaches are categorized into three categories which are as follows:

\begin{enumerate}
    \item Score-and-Search: In the score-and-search based approach, OAM algorithm requires the computation of the outlying degree of a query in each possible subspace in order to identify the subspace where it exhibits the highest degree of outlying characteristics w.r.t. the rest of the data.
    \item Feature Selection: In this approach, the problem of OAM is treated as a traditional problem of feature selection for classification. 
    \item Hybrid Approach: In the hybrid approach, the problem of OAM, is solved using a combination of score-and-search and feature selection based approach. 
\end{enumerate}

\section{Score-and-Search based approach} \label{sec:score-and-search}
To date, most of the studies that have been conducted to solve OAM problem belong to this category. The score-and-search approach required scoring function to measure the outlying degree of the given query. Then outlyingness of a query will be compared in all possible subspaces to detect the most outlying aspects. 

As far as we know, \cite{Zhang2004} is the earliest work, which addresses the problem of outlying aspect mining. Therein, the authors introduced a framework that detects the outlying subspace of a given query termed as HOS-Miner which stands for {\bf H}igh-dimensional {\bf O}utlying {\bf S}ubspace {\bf Miner}. They formulate the problem as: for a given data object, identify the subspaces in which this query object is considerably dissimilar or inconsistent w.r.t. the rest of the data objects. Moreover, this problem mathematically is stated as follows: for a given data object ${\bf q}$, find the set of subspaces $\mathcal{S}_\mathcal{F}$ such that for each subspace $S \in \mathcal{S}_F$, $OD_{S}({\bf q}) \geq \delta$, where $OD$ is the distance function (Equation \ref{EQ:HOS-Miner}), and $\delta$ is distance threshold. They described HOS-Miner as ``outlier $\rightarrow$ spaces" method.

In their work, they employed a distance-based scoring measure called {\bf O}utlying {\bf D}egree ($OD$ in short) to measure the outlyingness of the given query, which is the sum of the distances between the query and its k-nearest neighbors. The $OD$ of a query point ${\bf q}$ in subspace $S$ is calculated as :

\begin{equation} \label{EQ:HOS-Miner}
    OD_S({\bf q}) = \sum\limits_{x \in \aleph_S^k({\bf q)}} d_S(q,x)
\end{equation}

\begin{figure}[t]
\centering
    \includegraphics[scale=0.45]{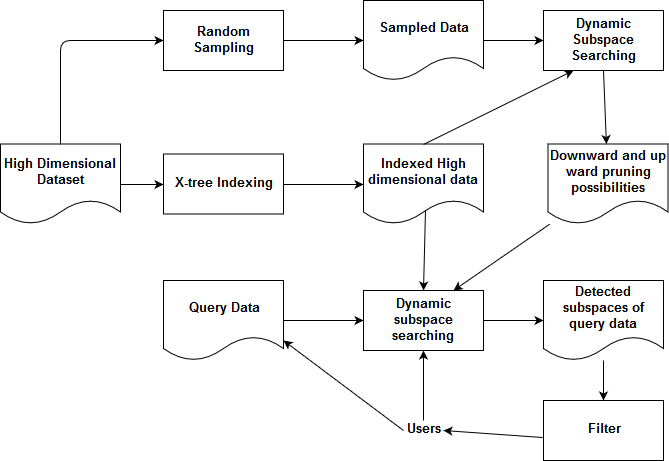}
    \caption{The overview of HOS-Miner \cite{Zhang2004}.}
    \label{fig:hos-miner}
\end{figure}

The process of HOS-Miner is shown in Fig. \ref{fig:hos-miner}. The proposed framework is divided into four steps. In the first step, the X-tree indexing module executes X-Tree \cite{Berchtold1996} indexing on the data set to enable $k$-nearest neighbor ($k$NN) search faster in subspace $S$. In the second step, the random sampling module randomly selects samples from the data set and then performs a dynamic subspace search to examine downward and upward subspace pruning possibilities of low to high dimensional subspaces. In the subsequent module, the subspace outlier detection module calculates the outlier score of the query and performs a dynamic subspace search to find subspaces where the query object deviates from the rest of the data. The last module is a filtering module, which filters out the most outlying subspace and returns to the user. 

Duan et al. (2015) \cite{Duan2015} introduce {\bf O}utlying {\bf A}spect {\bf Miner} ({\bf OAMiner} in short), which uses a Kernel Density Estimation (KDE) \cite{Silverman1986} based scoring measure to compute the outlyingness of query ${\bf q}$ in subspace $S$:

\begin{equation}
    f_{S}({\bf q}) = \frac{1}{n(2 \pi)^{\frac{m}{2}} \prod\limits_{i \in S} h_{i}} \sum\limits_{{\bf x} \in \mathcal{O}} e^ {- \sum\limits_{i\in S} \frac{(q.i - x.i)^2}{2 h^2_{i}}}
\end{equation}

\noindent where, $f_S({\bf q})$ is a kernel density estimation of ${\bf q}$ in subspace $S$, $m$ is the dimensionality of subspace $S$ ($|S|=m$), $h_{i}$ is the kernel bandwidth in dimension $i$. 

The study carried out by Duan et al. (2015)\cite{Duan2015}, stated that density is a bias towards high-dimensional subspaces -- density tends to decrease as dimension increases. Thus, to remove the effect of dimensionality biasedness, they proposed to use the density rank of the query as a measure of outlyingness. To find the most outlying subspace of query, the density of all data point needs to compute in each subspace, where the subspace with the best rank is selected as an outlying aspect of the given query.

OAMiner systematically enumerates all the possible subspaces. In OAMiner, the author has used the set enumeration tree approach \cite{Rymon1992}, which is widely used by the data mining research community. OAMiner searches for subspaces by traversing a depth-first manner \cite{Russell2009}. OAMiner used some anti-monotonicity properties to prune the subspaces. Given data set $\mathcal{O}$, a query object ${\bf q}$ and subspace $S$, if $rank(f_{S}({\bf q}))$ = 1, then every super-set of $S$ cannot be a minimal subspace and thus can be pruned.

OAMiner has two fundamental challenges:

\begin{enumerate}
    \item  OAMiner uses a density-based scoring function. Computing the density of each data point in each subspace is computationally expensive. Thus, it becomes infeasible in large and high dimensional data sets. The time complexity of finding the rank of $q$ in subspace $S$ is $O(n^2 m)$.
    \item OAMiner employs depth-first-search and utilizes anti-monotonicity property to prune subspace; therefore, an expensive search is required to find outlying aspects of the given query.
\end{enumerate}

The work of Vinh et al. (2016) \cite{Vinh2016} captures the concept of dimensionality unbiasedness and further investigates scoring functions, which is dimensionally unbiased. Dimensionality unbiasedness is an essential property for outlying measures because the query object is compared in different subspaces with a different number of dimensions. They proposed two novel outlying scoring metric (1) density $Z$-score and (2) \textbf{i}solation \textbf{Path} score (iPath in short). In their work, they showed that the proposed $Z$-score and iPath are dimensionally unbiased. 

Therein, the density $Z$-score is defined as follows:

\begin{equation}
   \mbox{Z-Score} (\tilde{f}_S({\bf q})) \triangleq \frac{\tilde{f}_S({\bf q}) -\mu_{\tilde{f}_S}}{\sigma_{\tilde{f}_S}}
\end{equation}

\noindent where $\mu_{f_S}$ and $\sigma_{f_S}$ are the mean and standard deviation of the density of all data instances in subspace $S$, respectively.

The iPath score is motivated by \textbf{i}solation \textbf{Forest} (iForest) anomaly detection approach \cite{Liu2008}. The intuition behind iForest is that anomalies are few and susceptible to isolation. iForest constructs $t$ trees, where each tree is constructed from randomly selected sub-samples $\psi$ ($\psi \ll n$). Later, it divides using the axis-parallel random splits. Since in the outlying aspect mining context, the main focus is on the path length of the query; thus, authors have ignored other parts of the tree. In outlying aspect mining, the intuition behind iPath score is that in the most outlying subspace, a given query is easy to isolate than the rest of the data.

The process of calculating the iPath of query ${\bf q}$ w.r.t. sub-samples $\psi$ of the data is 

\begin{equation}
    iPath_S({\bf q}) = \frac{1}{t} \sum\limits_{i=1}^t l_S^i({\bf q})
\end{equation}

\noindent where $l_S^i({\bf q})$ is path length of ${\bf q}$ in $i^{th}$ tree and subspace $S$.

\begin{figure}[t]
    \centering
    \includegraphics[scale=0.55]{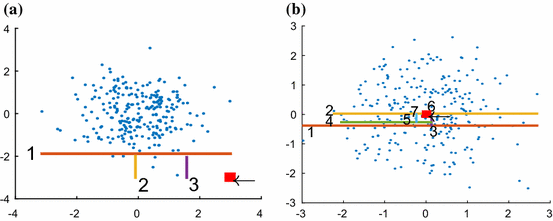}
    \caption{An illustrative example of iPath. The query is presented as red square. (a) A random isolation path of a query point where it is an outlier. (b) A random isolation path of a query point where it is an inlier \cite{Vinh2016}.}
    \label{fig:iPathDemo}
\end{figure}

The demo of iPath is presented in Fig. \ref{fig:iPathDemo}. In Fig. \ref{fig:iPathDemo}, the red square is a query point in $2$ dimensional space. Each horizontal or vertical numbered line represent splits. In Fig. \ref{fig:iPathDemo}(a), to isolate query, 3 splits are required, whereas 7 splits are required to isolate query in Fig. \ref{fig:iPathDemo}(b).

Vinh et al. (2016)\cite{Vinh2016} was the first to coin the term dimensionality unbiasedness.

\begin{definition}[\textbf{Dimensionality unbiased} \cite{Vinh2016}]
A dimensionality unbiased outlyingness measure ($OM$) is a measure of which the baseline value, i.e., average value for any data sample $\mathcal{O} = \{o_1, o_2, \cdots, o_n \}$ drawn from a uniform distribution, is a quantity independent of the dimension of the subspace S, i.e.,
    $$
        E[OM_S(x) | x \in \mathcal{O}] = \frac{1}{n} \sum\limits_{x \in \mathcal{O}} OM(x) = \mbox{const. w.r.t } |S|
    $$
    \label{def1}
    
\end{definition}

In \cite[Theorem 3]{Vinh2016}, it is proven that rank transformation and $Z$-score normalization have resulted in a constant average value in any data distribution. It is worth noting that the $Z$-score scoring function is not only normalized but also the variance of the normalized measures that are constant to dimensions.

The overall beam search process is divided into three stages. In the first stage, all $\mbox{1-D}$ subspaces are inspected to identify trivial outlying features. In the subsequent stage, an exhaustive search is performed on all possible $2$ dimensional subspaces. In the third stage, the beam search is implemented at level $l$. The beam algorithm only keeps top $W$ subspaces (that is called beam width) in the search process. The total number of subspace considered by beam algorithm is in the order of $O(D^2 + W \  \ D_{max})$ where $D_{max}$ is a maximum dimension of subspace, and $W$ is the beamwidth.

\begin{figure}[tb]
    \centering
    \includegraphics[scale=0.75]{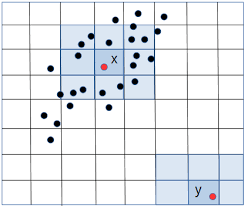}
    \caption{An illustrative example of the sGrid \cite{Wells2019}.}
    \label{fig:SGrid}
\end{figure}

\cite{Wells2019} introduced a simple grid-based density estimator called sGrid. sGrid is a smoothed variant of a grid-based density estimator \cite{Silverman1986}. Let $\mathcal{O}$ be a collection of $n$ data objects in $D$-dimensional space, $x.S$ be a projection of a data object $x \in \mathcal{O}$ in subspace $S$. The sGrid density of point $q$ is computed as points that fall in a bin that covers point $q$ and its surrounding neighbors. Fig. \ref{fig:SGrid} shows an illustrative example of sGrid, in which $x$ is estimated using $9$ bins while $y$ is estimated using $6$ bins. 

In their work, they showed that the proposed density estimator has advantages over the existing kernel density estimator in outlying aspect mining by replacing kernel density estimator to sGrid. By replacing KDE to sGrid density estimator, OAMiner \cite{Duan2015} and Beam \cite{Vinh2016} runs two orders of magnitude faster than their origin implementation. However, sGrid is not a dimensionally unbiased measure; hence it requires $Z$-Score normalization. Again, it makes sGrid computationally inefficient. 

\begin{figure}[t]
    \centering
    \subfloat[]{\includegraphics[width=0.4\textwidth]{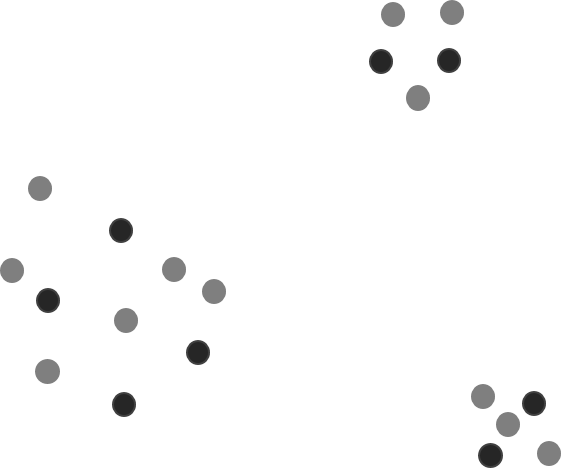}}\hspace{50pt}
    \subfloat[]{\includegraphics[width=0.4\textwidth]{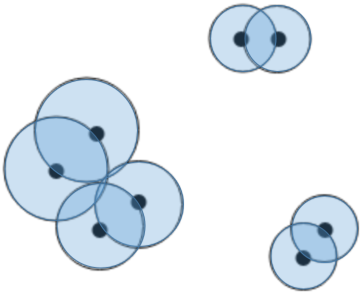}}
    \caption{(a) An example data set $\mathcal{O}$ (samples on dark black are selected to be in $\mathcal{D}_i$ to construct ${\mathcal M}_i$); and (b) Normal region is defined as the area covered by hyperspheres in ${\mathcal M}_i$ \cite{Samariya2020}.}
\label{fig:SiNNE-Model}
\end{figure}

Very recently, \cite{Samariya2020} proposed a \textbf{S}imple \textbf{I}solation score using \textbf{N}earest \textbf{N}eighbor \textbf{E}nsemble (SiNNE in short) measure which is motivated from Isolation using Nearest Neighbor Ensembles (iNNE in short) method for outlier detection \cite{Tharindu2017}. SiNNE constructs $t$ ensemble of models ($\mathcal{M}_1, \mathcal{M}_2, \cdots, \mathcal{M}_t$). Each model $\mathcal{M}_i$ is constructed from randomly chosen sub-samples ($\mathcal{D}_i \subset \mathcal{O}, |\mathcal{D}_i| = \psi < n)$. Each model have $\psi$ hyperspheres, where radius of hypersphere is the euclidean distance between $a$ ($a \in \mathcal{D}_i)$ to its nearest neighbor in $\mathcal{D}_i$. 
A working example of SiNNE model is constructed on $2$-Dimensional data set having 20 data objects and $\psi = 8$ presented in Fig. \ref{fig:SiNNE-Model}. The outlying score of ${\bf q}$ in model $\mathcal{M}_i$, $I(q|\mathcal{M}_i) = 0$ if $q$ falls in any of the ball and 1 otherwise. The final outlying score of ${\bf q}$ using $t$ models is :
\begin{equation}
    \mbox{SiNNE}(q) = \frac{1}{t} \sum\limits_{i=1}^t I(q|\mathcal{M}_i)
\end{equation}

In their work, they argue that $Z$-score normalization is biased towards a subspace having high-density variance and the definition of dimensionality unbiasedness is not sufficient. SiNNE is computationally faster than density and distance-based measures.

\paragraph{\textbf{Strengths and Weaknesses.}} The existing OAM score-and-search techniques show good performance. However, distance and density-based measures are computationally expensive. As a result, they are only applicable to very small data sets. The iPath score is a computationally fastest measure because it does not require any distance computation. However, the iPath score is unable to detect local outliers. sGrid density estimator is a great replacement of KDE density estimator because it is computationally efficient than KDE. However, sGrid is biased towards high dimensional subspaces. Thus, it requires $Z$-score normalization, which adds significant computational overhead. SiNNE is the second-fastest measure after iPath. However, iPath is unable to detect local outliers, whereas SiNNE can. In addition to that, it is an unbiased measure; hence there is no need for any normalization. The time complexity of each scoring measure is summarized in Table \ref{tab:features}.

\begin{table}[htb]
    \caption{The time complexity to compute the score of one query $q$ in a subspace by using different measures. Note that $n$ is the data size; $m$ is the dimensionality of subspace; $w$ is the block size in bitset operation, a parameter used in sGrid; $\psi$ is sub samples size and $t$ is number of sets, which are parameters used in iPath.}
    \label{tab:features}
    \centering
    \begin{tabular}{l @{\hspace{50pt}} l}
        \toprule\noalign{\smallskip}
        Scoring Measure & Time Complexity \\
        \noalign{\smallskip}\midrule\noalign{\smallskip}
        Density & $O(nm)$ \\
        Density Rank & $O(n^2m)$ \\
        Density $Z$-Score & $O(n^2m)$ \\
        iPath & $O(t \psi)$ \\
        sGrid $Z$-Score & $O(n^2m/w)$ \\
        SiNNE & $O(t \psi m + t \psi^2 m)$ \\
        \noalign{\smallskip}\bottomrule
    \end{tabular}
\end{table}

\section{Feature Selection} \label{sec:feature-selection}
Compared to the above mentioned approach, a little study is available for feature selection based methods. In the feature selection approach \cite{Micenkova2013, Dang2014}, firstly, the outlying aspect mining problem is transformed into classification and then performs some classical feature selection approaches to find an explanatory subspace of a given outlier.

In this line of work, \cite{Micenkova2013} is the earliest work which performs outlier explanation on the numeric data sets. They termed the outlying aspect mining problem as \textit{outlier explanation}. They formulate the problem of outlier explanation as: for a given outlier, detected by any outlier detection algorithm, find the possible explanation for that outlier.  In this work, authors assume that the outlier is given as a query (input) data, and their aim is to find an explanatory subspace. 

Outlier explanation converts problem of OAM in two class (inlier and outlier class) classification problem. For each outlier ${\bf q}$, a outlier class is generated from a Gaussian $\mathcal{N}_d(q, \Sigma)$, where $\Sigma$ is $D$ x $D$ scalar matrix and $\Sigma = \lambda^2 I$, and $\lambda$ = $\alpha \cdot \frac{1}{\sqrt{D}} \cdot d_k({\bf q})$, $d_k({\bf q})$ is the distance between ${\bf q}$ and its $k^{th}$ nearest neighbor. The negative class is constructed by $k$-nearest neighbors of outlier point ${\bf q}$ in full feature space and $k$ points from rest of the data set.

\cite{Angiulli2009} has studied the problem of outlier property detection and introduced the outlying property detection technique. Given a categorical data set, the goal is to find out the top $k$ set of attributes from which the query point ${\bf q}$ has the highest outliers score. \cite{Angiulli2017} proposed a version for the numeric data set. For a given data set $\mathcal{O}$ in $D$ dimensional space, a query object $q \in \mathcal{O}$ \cite{Angiulli2017} finds the pairs ($E$,$S$), satisfying $E \subseteq \mathcal{O}$ and $S \in D$ where $E$ is referred as explanation and $S$ referred as property (dimension). In 2014, \cite{Dang2014} introduced LOGP which stands for \textbf{L}ocal \textbf{O}utliers with \textbf{G}raph \textbf{P}rojection. LOGP is a novel technique that offers a solution to two problems, (1) outlier detection and (2) outlier interpretation.

\paragraph{\textbf{Strengths and Weaknesses.}} The advantages of these methods are that they do not require any subspace search, so these methods are faster than score-and-search methods. However, feature selection based methods depend on $k$ nearest-neighbor techniques. As pointed out in Vinh et al. (2016) \cite{Vinh2016}, $k$-nearest neighbors in full dimensional space is dramatically different from the $k$-NN in the subspace.

\section{Hybrid Approach} \label{sec:hybrid}
To the best of our knowledge, {\bf OARank} (stands for {\bf O}utlying {\bf A}spect Mining via Feature {\bf Ranking}) \cite{Vinh2015} is the only work which solves OAM problem using a hybrid approach. The proposed hybrid framework uses the strength of both feature selection based approach and score-and-search based approach. The OARank framework is a two-stage process. In the first stage, the OARank framework rank features as per their outlyingness of the query in that feature. In the second stage, the score-and-search technique is performed on the set of top-ranked features, where $m \leq D$. However, the second stage is optional. The top selected feature is either used for manual user inspection or user can perform score-and-search on top $k$ ranked features. 

The condition to choose $m$ features is as follows:
\begin{equation}
    SS = \min_{\substack{S \subset D \\ |S| = m}} \Bigg\{ C(m) \sum\limits_{i=1}^n \sum\limits_{\substack{t,j \in S \\ t<j}} K(q.j-o_{i}.j, h.j) K(q.t - o_{i}., h.t) \Bigg\}
\end{equation}

\noindent where $K(x-\mu,h) = (2\pi h^2)^{-\frac{1}{2}} \ 
\frac{\exp{-(x-\mu)^2}}{2h^2} $ is the one dimensional Gaussian kernel. $h$ and $\mu$ is bandwidth and center of Gaussian kernel respectively. $C(m) = \frac{2}{nm(m-1)2^{(m-2)}}$ is a normalization constant.

\paragraph{\textbf{Strengths and Weaknesses.}} The hybrid systems are built upon the connection between score-and-search and feature selection based approaches. OARank uses a kernel density estimator to determine subspace where it is minimized, which is again computationally prohibited in large and high dimensional data sets.

\section{Open Challenges} \label{sec:open-challenges}

Outlying aspect mining has slowly got little attention from researchers. However, there are many challenges that needs attention in the future. First and foremost challenge is that traditional outlying aspect mining score-and-search based approaches use distance or density estimation based scoring measures. These methods are easy to implement. However, these methods have a high time complexity, which is $O(n^2 D)$. Thus, they are infeasible in high dimensional and huge data sets. The most computationally expensive part of OAM is the computation of score, which is a repeated task for every data object in each subspace.

Another issue that still needs attention is that there is no such globally accepted evaluation measure for outlying aspect mining systems. Vinh et al. (2016) \cite{Vinh2016} proposed to use an entropy-based evaluation measure called consensus index in their work. However, Wells and Ting (2019) \cite{Wells2019} pointed out that, a consensus index is more suitable to evaluate clustering outcomes than assessing the outlierness of a query in a subspace. Therefore, one of the open research challenges is the development of an evaluation metric that can be used to evaluate detected outlying aspects of the given query by OAM systems. 

An important part of OAM is to search the subspaces, where a given data object is different from the rest of the data objects. By using systematic search methods, OAM has to compute outlierness of a given query in each subspace. This technique makes OAM methods computationally expensive. So an appropriate search technique is needed to reduce the effect of the curse of dimensionality. 

\section{Conclusion} \label{sec:Conclusion}
Outlying aspect mining is a new field, and a little is known about it among the research community, which motivates us to write this survey. In this survey, an attempt is made to summarise various ways in which the problem of outlying aspect mining has been solved in the past and discussed existing work, which is divided based on approaches. We have discussed the strengths and weaknesses of each approach in their respective categories. However, we are specifically interested in problems related to efficiency and effectiveness for high dimensional and large data sets. We believe there is still room for improvement in the area of outlying aspect mining, which offers lots of research opportunities in the future.

\subsubsection*{Acknowledgments}
This work is supported by Federation University Research Priority Area (RPA) scholarship, awarded to Durgesh Samariya. 

%
%
\bibliographystyle{splncs04}
\bibliography{reference}

\end{document}